\pdfoutput=1

\documentclass[11pt]{article}

\usepackage[final]{acl}

\usepackage{times}
\usepackage{latexsym}

\usepackage[T1]{fontenc}

\usepackage[utf8]{inputenc}

\usepackage{microtype}

\usepackage{inconsolata}

\usepackage{graphicx}

\usepackage{xcolor} 
\usepackage{textcase} 

\usepackage{booktabs}

\title{M-MiniGPT4: Multilingual VLLM Alignment via Translated Data}

\author{Seung Hun Han $^{\star\dagger}$\thanks{Corrsponding author: \texttt{eddieseunghunhan@gmail.com}} \\
  MBZUAI 
    \And
  Youssef Mohamed $^\star$\\
  KAUST 
  \\  
  \texttt{\{firstname.lastname\}@kaust.edu.sa}
  \\
    $^\star$ Equal Contribution
    ~ $^\dagger$ Work done while being an intern at KAUST 
    \And
  Mohamed Elhoseiny \\
  KAUST 
  }

\begin{document}
\maketitle

\begin{abstract}
This paper presents a Multilingual Vision Large Language Model, named M-MiniGPT4. Our model exhibits strong vision-language understanding (VLU) capabilities across 11 languages. We utilize a mixture of native multilingual and translated data to push the multilingual VLU performance of the MiniGPT4 architecture. In addition, we propose a multilingual alignment training stage that uses parallel text corpora to further enhance the multilingual capabilities of our model. M-MiniGPT4 achieves 36\% accuracy on the multilingual MMMU benchmark, outperforming state-of-the-art models in the same weight class, including foundation models released after the majority of this work was completed. We open-source our models, code, and translated datasets to facilitate future research in low-resource and multilingual settings.
\end{abstract}

\section{Introduction}
\label{sec:intro}

With the rise of powerful general-purpose Large Language Models (LLMs), multimodal extensions have begun to garner significant attention. In particular, Vision Large Language Models (VLLMs) combine the reasoning capabilities of LLMs with visual perception using a vision encoder. Notable early examples of open-source VLLMs include \cite{zhu2023minigpt4enhancingvisionlanguageunderstanding}, \cite{liu2023llava}, and \cite{dai2023instructblipgeneralpurposevisionlanguagemodels}. 

However, as early LLM development focused primarily on the English language, the derived VLLMs also tended to possess limited multilingual capabilities. Consequently, the benefiting audience has been restricted to English speakers, leaving approximately 75\% of the global population\footnote{cochrane.org/news/cochrane-evidence-different-languages} unable to benefit from advancements in these powerful open-source VLLMs.

Recently, open-source LLMs trained on multilingual data, such as Llama 3.1 \cite{dubey2024llama}, Qwen \cite{qwen2024qwen25technicalreport}, and Command R \cite{cohere2024}, have emerged. Similarly, Qwen-VL \cite{bai2023qwenvlversatilevisionlanguagemodel}, based on Qwen 2.5 \cite{yang2024qwen2}, has demonstrated improved multilingual capabilities. However, many of these models were not developed with multilinguality as their core objective, leading to limitations in language coverage. In this work, we explore the development and benchmarking of massively multilingual VLLMs using both synthetic and human-translated data. Furthermore, we demonstrate that the plug-and-play framework of MiniGPT4 is well-suited for multilingual learning, allowing it to scale with independent advancements in text-only LLMs and vision encoders.

VLLMs are typically trained in three stages. The first stage is designed to align the vision and language modalities and involves large-scale paired image-language data. Models following stage 1 tend to show weak reasoning performance; thus, a second stage involving high-quality instruction data is required to produce performant models. While data for both stages are readily available for English, high-quality multilingual multimodal data is scarce, and parallel multilingual multimodal datasets are virtually nonexistent. To mitigate this issue, we utilize state-of-the-art translation models to translate popular vision-language datasets. We show that translated data improves the model's multilingual performance without any notable degradation in English performance.

However, translated data does not account for the cultural and linguistic nuances that manifest only in natively collected datasets. As a result, relying solely on translation can result in sub-optimal multilingual VLLMs. To address this, we leverage parallel text corpora used for training machine translation models, as well as multilingual non-parallel text-only data, to improve the multilingual alignment of our models.

To summarize, our contributions are:
\begin{itemize}
    \setlength\itemsep{0em}
    \item We translate multiple vision-language datasets to create new multilingual resources.
    \item We demonstrate the use of parallel text corpora to improve the multilingual performance of VLLMs.
    \item We train a state-of-the-art (SOTA) multilingual VLLM based on the MiniGPT4 architecture.
    \item We translate the MMMU benchmark to assess the multilingual reasoning performance of VLLMs.
    \item We open-source all translated datasets and models to support the community.
\end{itemize}

\section{Related Work}
\label{sec:related}

\noindent\textbf{Large Language Models and Multilinguality.} LLMs have emerged as a transformative force in artificial intelligence, with success attributed to advances in GPU capabilities and large-scale training data. The field witnessed a paradigm shift with GPT-3~\cite{brown2020language}, demonstrating remarkable zero-shot capabilities. This sparked the development of numerous models, including open-source alternatives such as Bloom and OPT~\cite{scao2022bloom, zhang2022opt}, and proprietary models like Chinchilla~\cite{hoffmann2022training}, PaLM~\cite{chowdhery2022palm}, and Megatron-Turing NLG~\cite{smith2022using}. 

While LLaMA~\cite{touvron2023llama} introduced an approach with fewer parameters but more extensive training data, and the field continues to evolve with Llama 2/3, GPT-4, and Mistral~\cite{jiang2023mistral}, these models primarily focus on English. To address this, Multilingual Large Language Models (MLLMs) have emerged, excelling in cross-lingual transfer tasks. XLM-R~\cite{conneau2019unsupervised} pioneered cross-lingual capabilities, followed by models like Bloom and mT5~\cite{xue2020mt5} that intentionally incorporate substantial non-English data. Their instruction-tuned variants, Bloomz and mT0~\cite{muennighoff2022crosslingual}, have further advanced multilingual capabilities. Our research leverages these robust cross-lingual transfer capabilities to extend Vision and Language models into multilingual applications.

\noindent\textbf{VLLMs:} Recent advances in vision-language integration have focused on adapting LLMs to process visual information. Early approaches like VisualGPT~\cite{chen2022visualgpt} and Flamingo~\cite{alayrac2022flamingo} combined pre-trained LLMs with visual features. BLIP-2~\cite{li2023blip} introduced the Q-former to bridge visual and language representations. Building upon this, MiniGPT-4~\cite{zhu2023minigpt} enhanced performance by incorporating the Vicuna model. LLaVA~\cite{liu2023llava} aligned a frozen image encoder with LLaMA through instruction tuning, while InstructBLIP~\cite{dai2023instructblip} leveraged 26 diverse datasets. While effective for English, cross-lingual capabilities in these models remain largely unexplored.

Recently, models such as PALO \cite{PALO} have tackled the multilingual aspect of VLLMs, supporting visual reasoning for 10 languages via translated instruction datasets. Although PALO showed promising visual understanding, it performed poorly on visual reasoning benchmarks. We observed that PALO models excelled at lengthy descriptions but failed at direct question answering. In this paper, we show that this issue stems from the limited size of the PALO dataset. Accordingly, we provide a more diverse translated dataset, resulting in significantly better performance in both understanding and reasoning tasks.

\section{Datasets}
\label{sec:dataset}

VLLMs require high-quality multimodal data. We used No Language Left Behind (NLLB 1.3B) \cite{costa2022no} to translate popular vision-language datasets into 10 languages: Chinese, Hindi, Spanish, French, Arabic, Bengali, Russian, Urdu, Japanese, and Korean. Additionally, we translated the MMMU visual reasoning benchmark \cite{yue2024mmmu} using the same model. We utilize both V\&L datasets and text-only datasets.

\subsection{V\&L Datasets}

\begin{itemize}
    \item \textbf{Conceptual Captions} \cite{sharma2018conceptual}, \textbf{SBU} \cite{SBU}, and \textbf{LAION} \cite{schuhmann2021laion} are weakly-labelled datasets consisting of English image-caption pairs. They are used for Stage 1 pretraining to align vision and text modalities. Collectively, they consist of roughly 5 million instances.
    \item \textbf{LLaVA-Instruct} consists of 6.8 million English image-text pairs used to train the LLaVA 1.5 model \cite{liu2023llava}. The text consists of conversations and responses generated by GPT-4. We translated this dataset using NLLB to all 10 target languages. We refer to the translated version as \textbf{LAVAM}.
    \item \textbf{PALO} consists of 2 million image-text pairs used to train the PALO model \cite{PALO}. The dataset providers translated the LLaVA-665K dataset into 9 different languages.
    \item \textbf{Cambrian Image (CI)} is a high-quality multimodal dataset released for the Cambrian-1 family of models \cite{tong2024cambrian1}, consisting of approximately 58 million English image-text pairs. We translated this dataset to all 10 target languages and refer to the translated version as \textbf{CI M}.
    \item \textbf{WIT} is a natively multilingual dataset derived from Wikipedia, consisting of 130K image-article pairs \cite{WITSrinivasan_2021}. We filtered over 95\% of the original datapoints due to quality issues (corrupted text, stub articles). We used the NLLB-CLIP model \cite{visheratin2023nllbcliptrainperformant} to measure caption-image cosine similarity ($S_{I,C}$) and selected only pairs where $S_{I,C} \geq 0.0$.
\end{itemize}

\subsection{Text Corpora}
\begin{itemize}
    \item \textbf{Cambrian Text (CT)} is a high-quality text dataset released for the Cambrian-1 models \cite{tong2024cambrian1}, consisting of 22 million datapoints. We translated this to the target languages, naming the version \textbf{CT M}.
    \item \textbf{Flores} is a natively multilingual dataset derived from Wikipedia, consisting of 110K translation pairs between 11 languages \cite{nllb2022}.
    \item \textbf{XStoryCloze} is a human-translated paragraph completion dataset derived from StoryCloze \cite{xstorycloze}, consisting of 20K datapoints across 10 languages.
\end{itemize}

We combine Flores and XStoryCloze into \textbf{MText} in our experiments.

\begin{table}[t]
    \centering
    \begin{tabular}{c|c|c}
        \textbf{Model} & \textbf{E-MMMU} & \textbf{BT-MMMU} \\
        \hline
        Pretrained & 34.61   & 34.45 \\
        Finetuned & 34.14   & 33.02 \\
        \hline 
        \bottomrule
    \end{tabular}
    \caption{\textbf{Performance validation via back-translation.} We report the average accuracy after back-translation from all target languages. \textbf{E-MMMU}: English MMMU; \textbf{BT-MMMU}: Back-Translated MMMU.}
    \label{tab:back-translation}
\end{table}

\subsection{Multilingual MMMU Benchmark}
We utilized NLLB to translate the MMMU benchmark, which is designed to evaluate the reasoning capabilities of VLLMs. We validated the quality of our translation via evaluation before and after back-translation. specifically, we evaluated our model on the official English MMMU, translated MMMU to the target languages, and then back to English. Finally, we evaluated our model on the back-translated version. If the translation caused significant information loss, performance should drop; however, as shown in Table \ref{tab:back-translation}, performance remains consistent.

\section{Experiments} 
\label{sec:experiments}

\begin{table*}[t]
\centering
\begin{tabular}{l|l|cc}
\textbf{Stage 2} & \textbf{Stage 3} & \textbf{MMMU} & \textbf{MMMU Multi} \\
\hline
(ccSBU, LAION, LAVAM, PALO) & - & 31.02 & 29.83 \\
 \quad + CI & - & 34.14 & 31.21 \\
 \quad + CI M & - & 36.69 & \textbf{33.57} \\
 \quad + CI & CI M + CT M + MText & \underline{37.07} & 32.90 \\
 \quad + CI M & CI M + CT M + MText & \textbf{37.27} & \underline{33.45} \\
 \quad + CI & CI + CT & 35.19 & 32.93 \\
 \quad + CI & CI + CT + MText & 35.65 & 32.60 \\
\hline
\bottomrule
\end{tabular}
\caption{Ablation Studies on Training Data Combinations. \textbf{CI}: Cambrian Image; \textbf{CT}: Cambrian Text; \textbf{M}: Translated/Multilingual version.}
\label{tab:ablation}
\end{table*}

\subsection{Model Setup}
We base our model on the MiniGPT4 \cite{zhu2023minigpt} architecture. To support multilinguality, we replace the Vicuna LLM with Llama 3 \cite{dubey2024llama}, which demonstrates superior performance on multilingual tasks. Our training pipeline consists of three stages, each designed to enhance specific aspects of model performance.

\noindent\textbf{Stage 1} aims to align the visual and language modalities. We use large-scale image-caption datasets (Conceptual Captions, SBU, and LAION). Our experiments indicate that incorporating additional datasets at this stage does not yield performance improvements; thus, this stage remains consistent with the original MiniGPT4 implementation.

\noindent\textbf{Stage 2} enhances multilingual understanding by training with multilingual multimodal data. We leverage our translated datasets (ccSBU, LAION, LAVAM, PALO). We further experiment with the Cambrian Image (CI) dataset and its translated version (CI M).

\noindent\textbf{Stage 3} focuses on boosting multilingual capabilities. We conduct ablation studies using CI and Cambrian Text (CT) datasets in both original and translated versions (CI M, CT M), as well as the parallel corpora used for translation (MText).

\begin{table}[t]
\centering
\begin{tabular}{l|cc}
\textbf{Model}  & \textbf{MMMU} & \textbf{MMMU Multi} \\
\hline
PALO & 28.36 & 13.12 \\
Qwen-VL 2.5  & \textbf{52.89} & 25.46 \\
Our Model & 37.27 & \textbf{33.45} \\
\hline
\bottomrule
\end{tabular}
\caption{Comparison to SOTA Vision-Language Models.}
\label{tab:bench}
\end{table}

\subsection{Results}
Table \ref{tab:bench} compares our model with state-of-the-art vision–language models on the MMMU and MMMU Multi benchmarks. On MMMU Multi, our model substantially outperforms other fine-tuning approaches built on the same base model, improving from 13.12\% (PALO) to 33.45\%, despite both methods using the Llama-3 backbone. Our approach also exceeds the performance of the latest open-source foundational model, Qwen-VL 2.5, which achieves 25.46\% on this benchmark, highlighting the effectiveness of the proposed method for multi-modal, multi-step reasoning.

On the standard MMMU benchmark, Qwen-VL 2.5 attains higher accuracy (52.89\%) than our model. We attribute this gap primarily to differences in instruction tuning scale and data diversity, as Qwen-VL 2.5 benefits from more extensive instruction tuning than was applied in our setting.

Table \ref{tab:ablation} presents our ablation studies. Several key observations emerge:
\begin{itemize}
    \item Adding the Cambrian Image dataset (CI) in Stage 2 improves performance on both benchmarks.
    \item Using the translated version (CI M) in Stage 2 yields further improvements (36.69\% on MMMU and 33.57\% on MMMU Multi).
    \item The optimal configuration combines CI M in Stage 2 with CI M + CT M + MText in Stage 3.
    \item Including multilingual text data (MText) in Stage 3 generally improves performance when combined with translated datasets.
\end{itemize}

These results demonstrate the effectiveness of our three-stage training approach and the importance of incorporating multilingual multimodal data to enhance cross-lingual vision-language understanding.

\section{Conclusion}
\label{sec:conclusion}
In this paper, we presented M-MiniGPT4, a multilingual vision-language model that demonstrates strong performance across 11 languages. Our approach leverages a three-stage training process that effectively combines native multilingual data with translated datasets to optimize cross-lingual vision-language understanding. We demonstrated that using translated vision-language data significantly improves multilingual performance and that incorporating parallel text corpora further enhances the model's capabilities.

Our experiments show that M-MiniGPT4 achieves state-of-the-art multilingual performance on the MMMU Multi benchmark (33.45\%), substantially outperforming existing models like Qwen-VL 2.5 and PALO in multilingual visual reasoning tasks. By open-sourcing our models and translated datasets, we facilitate further research in multilingual multimodal AI, making these technologies more accessible to non-English speakers worldwide.

\section{Limitations}
Despite promising results, M-MiniGPT4 faces several limitations:
\begin{itemize}
    \item \textbf{Translation Nuance:} Reliance on machine translation may not fully capture cultural nuances and linguistic subtleties present in natively collected multilingual data.
    \item \textbf{Language Coverage:} While our model was finetuned on 11 languages, this covers only a fraction of the world's languages.
    \item \textbf{Resource Disparity:} Translation quality varies, with high-resource languages (e.g., Spanish, French) benefiting from better translations compared to lower-resource languages (e.g., Bengali, Urdu).
    \item \textbf{Evaluation:} Our metrics may not comprehensively assess all aspects of cross-cultural understanding in visual reasoning.
    \item \textbf{Inherited Bias:} Reliance on pretrained LLMs inherits the biases and limitations inherent in the base models.
\end{itemize}

Future work should focus on expanding language coverage, incorporating more natively collected multilingual data, and developing nuanced evaluation frameworks for cross-cultural understanding.

\bibliography{custom}





\end{document}